\documentclass[conference]{IEEEtran}
\IEEEoverridecommandlockouts
\usepackage{cite}
\usepackage{amsmath,amssymb,amsfonts}
\usepackage{graphicx}
\usepackage{textcomp}
\usepackage{xcolor}

\usepackage{times}

\usepackage{soul}
\usepackage{url}
\usepackage{hyperref}
\usepackage[utf8]{inputenc}
\usepackage{graphicx}

\usepackage{booktabs}
\usepackage{tabulary,booktabs}

\newcommand{\name}{{\texttt{FedTriNet}}}

\usepackage{CJK}
\usepackage{algorithm}
\usepackage{algorithmicx}
\usepackage{algpseudocode}

\floatname{algorithm}{Algorithm}

\def\BibTeX{{\rm B\kern-.05em{\sc i\kern-.025em b}\kern-.08em
    T\kern-.1667em\lower.7ex\hbox{E}\kern-.125emX}}
\begin{document}

\title{FedTriNet: A Pseudo Labeling Method with Three Players for Federated Semi-supervised Learning\\

}
% \IEEEoverridecommandlockouts
% \IEEEpubid{\makebox[\columnwidth]{978-1-6654-3902-2/21/\$31.00~
% \copyright~2021 IEEE \hfill} \hspace{\columnsep}\makebox[\columnwidth]{}} 

\makeatletter
\def\ps@IEEEtitlepagestyle{%
  \def\@oddfoot{\mycopyrightnotice}%
  \def\@evenfoot{}%
}
\def\mycopyrightnotice{%
  {\hfill \footnotesize 978-1-6654-3902-2/21/\$31.00~
\copyright~2021 IEEE\hfill}
}
\makeatother

\author{\IEEEauthorblockN{Liwei Che}
\IEEEauthorblockA{\textit{College of IST} \\
\textit{Pennsylvania State University}\\
State College, USA \\
lwche@psu.edu}
\and
\IEEEauthorblockN{Zewei Long}
\IEEEauthorblockA{\textit{Department of Computer Science} \\
\textit{University of Illinois Urbana-Champaign}\\
Champaign, USA \\
zeweil2@illinois.edu}
\and
\IEEEauthorblockN{Jiaqi Wang}
\IEEEauthorblockA{\textit{College of IST} \\
\textit{Pennsylvania State University}\\
State College, USA \\\
jqwang@psu.edu}
\and
\IEEEauthorblockN{Yaqing Wang}
\IEEEauthorblockA{\textit{School of Electrical and Computer Engineering} \\
\textit{Purdue University}\\
West Lafayette, USA \\
wang5075@purdue.edu}
\and
\IEEEauthorblockN{Houping Xiao}
\IEEEauthorblockA{\textit{Institute for Insight} \\
\textit{Georgia State University}\\
Atlanta, USA \\
hxiao@gsu.edu}
\and
\IEEEauthorblockN{Fenglong Ma}
\IEEEauthorblockA{\textit{College of IST} \\
\textit{Pennsylvania State University}\\
State College, USA \\
fenglong@psu.edu}
}

\maketitle

\begin{abstract}
Federated Learning has shown great potentials for the distributed data utilization and privacy protection. Most existing federated learning approaches focus on the supervised setting, which means all the data stored in each client has labels. However, in real-world applications, the client data are impossible to be fully labeled. Thus, how to exploit the unlabeled data should be a new challenge for federated learning. Although a few studies are attempting to overcome this challenge, they may suffer from information leakage or misleading information usage problems. To tackle these issues, in this paper, we propose a novel federated semi-supervised learning method named {\name}, which consists of two learning phases. In the first phase, we pre-train {\name} using labeled data with FedAvg. In the second phase, we aim to make most of the unlabeled data to help model learning. In particular, we propose to use three networks and a dynamic quality control mechanism to generate high-quality pseudo labels for unlabeled data, which are added to the training set. Finally, {\name} uses the new training set to retrain the model. Experimental results on three publicly available datasets show that the proposed {\name} outperforms state-of-the-art baselines under both IID and Non-IID settings.
\end{abstract}

\begin{IEEEkeywords}
federated learning, semi-supervised learning, pseudo labeling
\end{IEEEkeywords}

\section{Introduction}
Federated learning~\cite{McMahan2017CommunicationEfficientLO, li2018deeper, Chen2019FederatedLO} has furnished a concrete solution to the training of machine learning models among decentralized data deployment networks with relative stable privacy preservation. A central server helps multiple clients collaborate on learning a global model, which outperforms any local models. This distributed framework contributes a series of advantages to the protection of data privacy, access rights, and security. 

However, several practical issues still shackle federated learning aggregation and affect its performance. For instance, the clients tend to generate a large amount of data, but they lack labels or only contain a few labels. While existing federated methods such as FedAvg~\cite{McMahan2017CommunicationEfficientLO} mainly focus on the supervised scenario where client data are fully labeled. It is crucial to get full access to the information included inside the unlabeled data to improve the global model performance.

Only a few studies are considering the unlabeled data, such as FedMatch~\cite{Jeong2020FederatedSL} and FedSem~\cite{Albaseer2020ExploitingUD}. 
FedMatch~\cite{Jeong2020FederatedSL} introduces the inter-client consistency loss and additive parameter decomposition to disjointly learn on both labeled and unlabeled data. However, as this approach needs to collect information from neighboring clients, it may leak sensitive information.
FedSem~\cite{Albaseer2020ExploitingUD} uses a simple two-phase pseudo-labeling based method for semi-supervised learning applications. If the performance of the first phase, i.e., pretraining the model with labeled data, is poor, it would introduce error messages in the subsequent marking process of pseudo labels, which seriously affects the learning effect. 

To address those problems, in this paper, we propose a novel two-phase learning framework named {\name} to guarantee information privacy and automatically generate high-quality pseudo labels via three networks. In the first phase, we pretrain the framework using labeled data on each client with FedAvg~\cite{McMahan2017CommunicationEfficientLO} like FedSem~\cite{Albaseer2020ExploitingUD}. In the second phase, we aim to generate high-quality pseudo labels for unlabeled data and further use them for retraining each local model. Towards this end, we design a new approach by considering three client networks and automatically generating a threshold as the criteria to filter out low-quality pseudo labels in each client. 

\textbf{Designing of Three Players.} In particular, the first network is the client model trained with labeled data in each client, which has good classification ability. In a deep neural network, such ability is usually determined by the last few layers.
The second one is the global model aggregated by all the client models, which usually has a strong ability to extract features using the first few layers.
The third model combines the client model and the global model, which tries to unify both models' advantages by borrowing the first few layers from the global model and the last few layers from the client model. Then, the combined model conducts finetuning with the labeled client data.  

\textbf{Pseudo Label Generation.} In each client, {\name} runs three networks on unlabeled data to output three prediction probability vectors, which are further used to generate the pseudo labels for unlabeled data. To guarantee the quality of the pseudo labels, we design a dynamical control mechanism to generate a global-level threshold $\theta$. Remarkably, each client will identify the maximum probability value and then upload it to the server. The server will average the uploaded client-level maximum probability values and distribute the maximum value, i.e., $\theta$, to each client. To carefully add the pseudo labeled data to the training set, a dynamic control mechanism is designed to make that $\theta$ decreases with the increase of the number of global training rounds. If the maximum probability value of the three prediction probability vectors is larger than $\theta$, then the corresponding unlabeled data will be added to the training data.

Finally, {\name} will retrain the client model using both the real labeled data and the pseudo-labeled data. Note that since there are three models in each client, we choose to retrain the finetuned combined model, which is significantly different from FedAvg~\cite{McMahan2017CommunicationEfficientLO} and FedSem~\cite{Albaseer2020ExploitingUD}. 
We evaluate the proposed {\name} on three benchmark image datasets under both IID and Non-IID data distribution settings compared with state-of-art baselines. Experimental results show the effectiveness of the proposed {\name} framework.

The remainder of this paper is organized as follows. Section~\ref{sec:related_work} systematically reviews the recent related work. Section~\ref{sec:framework} introduces the details of the proposed {\name}. Section \ref{sec:experiemts} presents experimental setups, results and analysis compared with baselines. Section~\ref{sec:con} concludes.
\section{Related Work}\label{sec:related_work}
This section systematically reviews the studies on federated learning, federated semi-supervised learning, and semi-supervised learning.

\subsection{Federated Supervised Learning}
Federated learning provides an efficient and privacy-preserved collaboration strategy for mutually training between different data owners, such as distributed data centers, customers and diverse institutions. The majority of federated learning works focus more on supervised learning scenarios and solving three challenges: statistical heterogeneity~\cite{DBLP:journals/corr/abs-1806-00582,DBLP:journals/corr/abs-1812-06127,DBLP:journals/corr/abs-1811-12629}, system constraints~\cite{Caldas2018ExpandingTR,Wang2019CMFLMC,Chen2018FederatedMW}, and trustworthiness~\cite{Bhowmick2018ProtectionAR,Geyer2017DifferentiallyPF,Bonawitz2016PracticalSA}. 
%In this paper we mainly focus on the statistical heterogeneity challenge on semi-supervised learning scenario in both IID and Non-IId setting, aiming to achieve better model performance on classification.
% Previous work has made a series of attempts to tackle this challenge from different angles. 
In particular, \cite{DBLP:journals/corr/abs-1806-00582} uses a shared server-stored dataset to help the clients achieve higher performance in Non-IID settings; \cite{DBLP:journals/corr/abs-1811-12629} applies adjustment on the SGD convergence of federated learning; and \cite{DBLP:journals/corr/abs-1812-06127} adds regularization terms on the loss function during the local training process on the clients to constrain the divergence between the global model and local ones.

\subsection{Federated Semi-supervised Learning}
A more realistic setting in federated learning is federated semi-supervised learning, i.e., simultaneously considering both labeled and unlabeled data.  However, the introduction of the unlabeled data will significantly increase the difficulty of the problem. The studies on federated semi-supervised learning are still at the baby step, but more and more researchers are paying attention to this research topic. In \cite{Albaseer2020ExploitingUD}, the authors propose a simple two-phase pseudo-labeling based method for semi-supervised learning application, and in  \cite{Jeong2020FederatedSL}, the authors introduce the inter-client consistency loss and additive parameter decomposition to disjointly learn on both the labeled and unlabeled data.
However, the existing methods are efficient while may violate the clients' privacy or have poor performance with scarce labeled data, which are severe disadvantages for a federated semi-supervised learning problem. 
\subsection{Semi-supervised Learning}
Semi-supervised learning {SSL} is a research field of practical significance and value to extract effective information from unlabeled data and help the model achieve better training effect and performance~\cite{4787647}. 
The previous SSL work shows a series of diverse and coherent solutions. An intuitive approach is pseudo labeling, which uses the pretrained model to label the unlabeled data. \cite{Lee2013PseudoLabelT} introduces a dynamic decision threshold to help the model labeling the data.  Another effective and well-known strategy is to add consistency regularization on the training loss~\cite{Rasmus2015SemisupervisedLW,Tarvainen2017MeanTA,Laine2017TemporalEF,Miyato2019VirtualAT,park2017adversarial}. \cite{NIPS2017_86e78499} presents an SSL method based on three neural networks, which characterize the conditional distributions between images and labels. In \cite{Athiwaratkun2019ThereAM}, the authors suggest that the flat platform of SGD leads to the convergence dilemma of consistency-based SSL. UDA~\cite{Xie2019UnsupervisedDA}, ReMixMatch~\cite{Berthelot2019ReMixMatchSL}, and Fixmatch~\cite{Sohn2020FixMatchSS} mix plenty of practical methods and do further exploration. \cite{cascantebonilla2020curriculum} adapts curriculum learning idea into pseudo label method with self-training strategy, especially for the setting with a small set of labeled data and a large set of unlabeled data.
\section{FedTriNet Framework}\label{sec:framework}
The goal of federated semi-supervised learning is to learn a global model $G$ via collaboratively training $K$ local client models $\mathcal{L} = \{l^k\}_{k=1}^K$. In this paper, we focus on the following setting: Each client stores both labeled data $\mathcal{D}_L^k = \{ (\mathbf{x}^k_i,y^k_i) \}_{i=1}^{N_L^k}$ and unlabeled data $\mathcal{D}_U^k = \{ \mathbf{x}^k_j \}_{j=1}^{N_U^k}$, where $y^k_i \in \{1, \cdots, M\}$ is the corresponding label of the data instance $\mathbf{x}^k_i$, $N_L^k$ denotes the number of labeled data of the $k$-th client, and $N_U^k$ denotes the number of unlabeled data of the $k$-th client . Note that there are no data at the server side. To learn the global model $G$, we design a simple yet effective framework named {\name}. Next, we present the details of our framework.

% Aiming to learning an intensified  global model without violation of the data privacy, \textbf{Federated Learning} provides an efficient collaboration learning strategy for distributed data distribution scenario. This section will provide the introduction of the framework and working mechanism of our proposed method {\verb|FedTriNet|}.

% Let $G$ denotes the global model and $\mathcal{L} = \{l^k\}_{k=1}^N$ denote a set of local models for $N$ client\textcolor{blue}{}s. For the $k$-th client, $\mathcal{D}_L^k = \{ (\mathbf{x}^k_1,y^k_1),\cdots, (\mathbf{x}^k_n,y^k_n) \}$ represents a set of labeled data, and $\mathcal{D}_U^k = \{ (\mathbf{x}^k_1,-1),\cdots, (\mathbf{x}^k_n,-1) \}$ for unlabeled one, where $\mathbf{x}^k_i$ ($i \in\{1, \cdots, n\}$) is a data instance, $y^k_i \in \{1, \cdots, M\}$ is the corresponding label which for unlabeled data is -1 initially, and $M$ is the number of label categories. The total communication round $T = T_1 + T_2$, where $T_1$ is the global training rounds number for Pre-training stage and $T_2$ for Semi-supervised training stage.

\begin{figure*}[htbp]  
\centering  
\includegraphics[height=6cm, width=14cm]{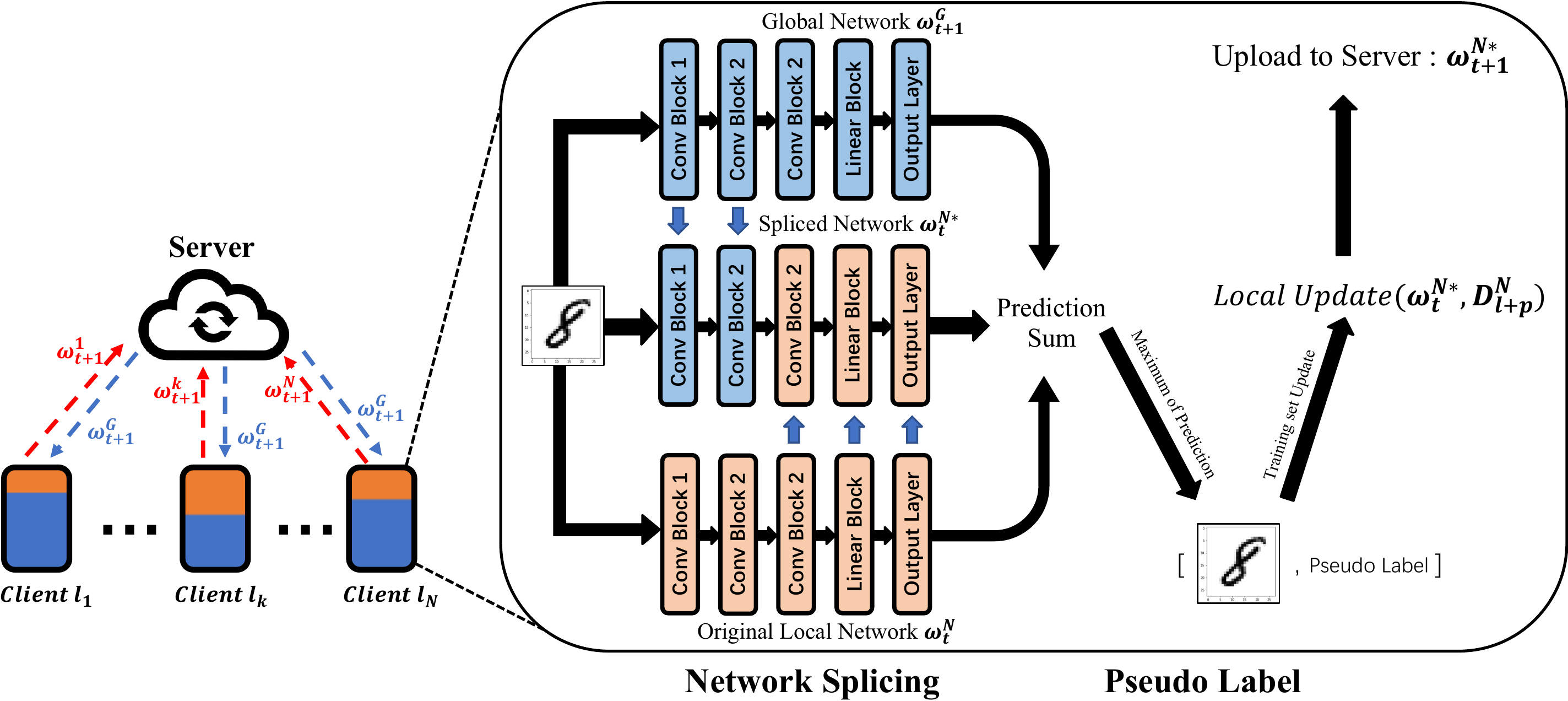}  
\caption{The proposed {\name} Framework} 
\label{Fig:overview}
\end{figure*}

\subsection{Model Overview}
Figure~\ref{Fig:overview} shows the flow of the proposed framework {\name}. {\name} consists of two modules, i.e., \emph{local training} and \emph{server update}. In the local training module, each client $k$ trains a local model $l^{k}$ using both labeled and unlabeled data. The parameters of $l^k$, which is denoted as $\boldsymbol{\omega}^k$, will be uploaded to the server. In the server update module, the server learns a global model $G$ by aggregating $K$ randomly uploaded local models, i.e., 
\begin{equation}
\label{server_update}
\boldsymbol{\omega}^G=\sum_{k=1}^{K} \frac{\boldsymbol{\omega}^k}{K},
\end{equation} 
where $\boldsymbol{\omega}^G$ is the the parameters of $G$.
$\boldsymbol{\omega}^G$ will be then distributed to each local client. This procedure will be repeatedly executed until the global model $G$ converges.  

In particular, the local training module of the proposed {\name} has two stages, which are \emph{pre-training} and \emph{pseudo label learning}. The goal of the pre-training stage is to train each local model and global model $T_1$ rounds only using labeled data. Then in the pseudo label learning stage, {\name} generates a pseudo label for each unlabeled data using three networks, which are original local network, downloaded global network, and a spliced network separately. 

The spliced network is a combination of the original local network and global network. Here, we assume that the global model's low-level feature extraction ability is better than that of the local model, which can be represented by the first $n$ layers of $G$. However, the local model can capture the classification characteristics of local data, which can be described by the last $m$ layers of $\boldsymbol{\omega}^k$. Thus, we can obtain a new network with $n+m$ layers to predict a pseudo label for each unlabeled sample.

By aggregating the outputs of the three networks, we can finally assign labels to unlabeled data. Using both labeled and pseudo labeled data, we can run a local training module to update the parameters, which will be uploaded to the server to update the global parameters. The new global parameters will also be distributed to each local client until they converge or the procedure runs $T_2$ rounds. 
The server update uses Eq.~(\ref{server_update}), and next, we will present the details of the local training in the proposed {\name} framework.

% Common \textbf{Federated Learning} usually uses two stages as one communication round,\textbf{ Local Training} and \textbf{Server Update}. Here in our model, above communication round, we divide the training process into two stages, the \textbf{Pre-training stage} and \textbf{Semi-supervised training stage}.

% During the first $T_1$ rounds, the whole federated system will execute supervised training based on the local labeled dataset ${D}_L^k$, which is also the initial training set. 

% In the second phase, semi-supervised training achieved by a novel pseudo-labeling method with three different models for each client based on consensus mechanism. The pseudo-label process happens on every client who participates in the global aggregation. 
% \begin{equation}
%     Prediction_{Sum}(x_u) = C_{old}(x_u)+C_{new}(x_u)+C_{G}(x_u)
% \end{equation}
% \begin{equation}
%     Pseudo Label = Max index(Prediction_{Sum}(x_u))
% \end{equation}

% Once the pseudo label is decided and the probability prediction is greater than a dynamic threshold value, this unlabeled data would be seen as a pseudo-labeled data with high confidence and added into the training set for the following local training epochs. This procedure would repeat until it reached the pre-set training round $T_2$. 

\subsection{Pre-training Stage}
The proposed {\name} framework aims to generate pseudo labels for unlabeled data and then to update the local model using both labeled and pseudo labeled data. The critical issue of this approach is how to guarantee the quality of the generated pseudo labels. Towards this end, we propose to pre-train the local and global models only using labeled data by optimizing the following loss function as FedAvg~\cite{McMahan2017CommunicationEfficientLO}:
\begin{equation}\label{eq:fedavg}
L_l(\mathcal{D}_L^k) = \min \left[\frac{1}{N_L^k} \sum_{i=1}^{N_L^k} CE \left(f\left(\mathbf{x}^k_i; \boldsymbol{\omega}^k\right), y^k_i\right)\right],
\end{equation}
where $L_l(\mathcal{D}_L^k)$ denotes the total loss, $CE$ is the cross-entropy loss, $f(\cdot;\cdot)$ represents the neural network such as convolutional neural network (CNN), and $\boldsymbol{\omega}^k$ is the parameter set. Then Eq.~(\ref{server_update}) is used to obtain the parameter set $\boldsymbol{\omega}^G_t$ of the global model $G$. 

In each communication round, the clients will download the global model's parameter for local training with predefined epochs from the server. After that, part of the clients will take part in the global aggregation that their local model parameters will be uploaded to the server. We repeatedly run this procedure $T_1$ times to pretrain both local and global models, and then $\name$ starts to consider the unlabeled data..

\subsection{Pseudo Label Learning Stage}
To make fully use of unlabeled data, a straightforward approach is to generate pseudo labels based on the pre-trained model in the pre-training stage. However, there are two kinds of models for each client, i.e., a local model $l^k$ and a global model $G$. The local model may perform better when the unlabeled data follow a similar distribution as the labeled data. However, real-world applications may not satisfy this constraint. The global model $G$ is aggregated by several local models. Using $G$ to generate the pseudo labels may not capture the characteristics of local models. Thus, either using local models or the global model may be prone to generate incorrect labels, further introducing incorrect information to model learning.

To guarantee the quality of pseudo labels as much as possible, in this paper, we introduce a combined model for each client, a combination of each local model and the global model. Intuitively, the shallow layers of deep neural networks focus more on low-dimensional feature learning, which can be shared even for different images. On the contrary, the class-related features of an image are abstracted into deeper layers, which are uniqueness. Based on this intuition, we can assemble a new network using the shallow layers' parameters of the global network that have better generalization ability and deep layers of the local network for capturing class-specific characteristics.

For instance, a convolutional neural network consists of three convolutional layers and two full connection layers. We usually select the parameters of the first two convolutional layers of the global network and the parameters of the full connection layers of the local network to form a new combined network. Note that the specific method of interception and the selection of layers are influenced by data type, network structures, and training parameter settings. 

% In the pseudo-label process, the pseudo label of one unlabeled data is decided by a mutual output based on the sum of the predictions of three players. Different from the majority voting strategy, which uses one-hot coding to adapt the position with the highest vote identified as the category to which the input belongs, our method uses softmax output where the location of the output to which the input belongs is a probability predictions that add up to 1. This could avoid some statistic error arisen in the decision process, such as three different votes, rounding error，etc.

% In this work, we used the intercepted layer number as a hyper-parameter to investigate in the subsequent ablation experiment section. Based on this three players' game, we also set a dynamic threshold, which will help the pseudo-labeling framework to maintain and update the local train set for each client.

\subsubsection{Multi-view Pseudo-labeling}
In the pseudo-labeling process, the pseudo label of one unlabeled data is decided by a mutual output based on the sum of the prediction probabilities of three players. Different from the majority voting strategy, which uses one-hot coding to adapt the position with the highest vote identified as the category to which the input belongs, our method uses the outputs of the softmax layer, where the location of the output to which the input belongs is a probability value. This could avoid some statistical errors arisen in the decision process, such as three different votes, rounding errors. 

Let $\mathbf{p}_{l^k}(\mathbf{x}_j^k)$ be the probability vector predicted by the local models $l^k$ with parameters $\boldsymbol{\omega}^k$ on the unlabeled data $\mathbf{x}_j^k$, and $\mathbf{p}_{G}(\mathbf{x}_j^k)$ be the probability vector outputted by the global model $G$. Let $c^k$ denote the combined model and  $\mathbf{p}_{c^k}(\mathbf{x}_j^k)$ be the outputted probability vector. 
Note that in our implementation, \emph{we use labeled data to fine-tune the model $c^k$ first and then use it to make predictions}.
Thus, the pseudo label of the unlabeled data $\mathbf{x}_j^k$ is
\begin{equation}\label{eq:triple_prediction}
\begin{split}
    \mathbf{p}_j^k &= \frac{1}{3}\left[\mathbf{p}_{l^k}(\mathbf{x}_j^k) + \mathbf{p}_{G}(\mathbf{x}_j^k) + \mathbf{p}_{c^k}(\mathbf{x}_j^k)\right],\\
    \hat{y}_j^k &= \operatorname*{arg\,max} \mathbf{p}_j^k.
\end{split}
    % \hat{y}_j^k = \operatorname*{arg\,max} \left[\frac{1}{3}\left(\mathbf{p}_{l^k}(\mathbf{x}_j^k) + \mathbf{p}_{G}(\mathbf{x}_j^k) + \mathbf{p}_{c^k}(\mathbf{x}_j^k)\right)\right].
\end{equation}

In order to use the pseudo labeled data to update the model, we must guarantee the quality of the pseudo labels. In other words, we cannot directly use all the pseudo labeled data and only use the data with high confidence. Thus, we design the following mechanism to control the quality of pseudo-labeled data dynamically. In particular, {\name} dynamically generates a global threshold $\theta$. If the maximum probability of unlabeled data is greater than $\theta$, then the corresponding data will be added to the training set. Next, we will how to estimate the value of $\theta$.

\subsubsection{Dynamic Pseudo-labeled Data Selection}
Towards the goal of generating a global threshold $\theta$, we first run the global model $G$ on each unlabeled sample $\mathbf{x}_j^k$ stored in each client $k \in \{1, \cdots, K\}$ to obtain the prediction $\mathbf{p}_{G}(\mathbf{x}_j^k)$. Then we can have the maximum probability of $\mathbf{p}_{G}(\mathbf{x}_j^k)$, i.e., $max(\mathbf{p}_{G}(\mathbf{x}_j^k))$.
Since there are $N_U^k$ unlabeled data in client $k$, we can obtain $N_U^k$ maximum probability values, i.e., $\{max(\mathbf{p}_{G}(\mathbf{x}_j^k))\}_{j=1}^{N_U^k}$. Finally, the maximum predictive probability of all the unlabeled data is 
\begin{equation}
    \theta^k = max\{max(\mathbf{p}_{G}(\mathbf{x}_1^k)), \cdots, max(\mathbf{p}_{G}(\mathbf{x}_{N_U^k}^k))\}.
\end{equation}

Since there are $K$ clients, for each client, we can obtain a client-level threshold. These $K$ thresholds are uploaded to the server to generate the global-level threshold $\theta$ as follows:
\begin{equation}
    \theta(t)=
\begin{cases}
\alpha \bar{\theta} & \text{$t < 10$},\\
\frac{(100-2t)}{100} \alpha \bar{\theta} & \text{$10\leq t <35$},\\
\frac{1}{2} \alpha \bar{\theta} & \text{ $t \geq 35$}, 
\end{cases}
\label{eq:threshold}
\end{equation}
where $t$ represents the number of communication rounds in the pseudo label learning stage, $\alpha$ is a predefined hyper-parameter to control the threshold, and $\bar{\theta}$ denotes the average of all the uploaded client-level thresholds, i.e., $\bar{\theta} = \frac{1}{K}\sum_{k=1}^K \theta^k$.
% after the mutual labeling stage, the prediction value of each unlabeled data will get compared with a dynamic threshold value, which is defined as follows:
% \begin{equation}\label{eq:threshold}
%     \theta_t = \frac{\alpha_t}{U} \bigl\{\sum_{k=1}^{U} \operatorname*{max} \left[\mathbf{p}_{l^k}(\mathbf{x}_j^k) + \mathbf{p}_{G}(\mathbf{x}_j^k) + \mathbf{p}_{c^k}(\mathbf{x}_j^k)\right] \bigr\}
% \end{equation}
% \textcolor{red}{this eq is still not clear. All the clients use one threshold? or each client has a threshold? what do you mean "the average maximum prediction value"?}
% where $\alpha_t$ is the dynamic damping factor, which is initialized by $1.1$ and then will decay 3\% per training round. $U$ is the total number of the unlabeled data. \textcolor{red}{data of each client?}
The motivation behind Eq.~(\ref{eq:threshold}) is that we want the local model to be more stable in the first few rounds of the pseudo label process. In order to avoid updating too many pseudo labeled data into the training set at one time, a larger threshold is used at the beginning of the pseudo label learning stage (i.e., $t < 10$) by setting $\alpha = 0.93$ (experimental result) in the experiment. In such a way, only a tiny amount of high-quality pseudo labeled data will be added to the training first. With the increase of the communication rounds, the threshold value will decrease. In other words, there will be more data to be added to the training set.

The global threshold $\theta(t)$ using Eq.~(\ref{eq:threshold}) is then distributed to each client $k$. If $max(\mathbf{p}_j^k)$ in Eq.~(\ref{eq:triple_prediction}) is greater than $\theta(t)$, then the corresponding unlabeled data will be added to the training set. Let $\mathcal{D}_{P}^k$ denote the selected pseudo labeled data, which will be used to retrain the local model.

% initialized by $\alpha_t = 1.1$ times the average maximum prediction value. This because in the first few rounds of the pseudo label process, we wanted the local model to be more stable. In order to avoid updating too much pseudo labeled data into the training set at one time, a higher initial threshold was set by $10\%$ higher than the average prediction value, and a small amount of high-quality pseudo labeled data was added to the training first  When the prediction value is greater than the threshold, this pseudo labeled data will be added into the train set for next round local training. The threshold value will decay along with the training round by 3\% per round.

% \begin{equation}
%     Threshold = \frac{\alpha_t}{U} \bigl\{\sum_{k=1}^{U} \operatorname*{max} \left[\mathbf{p}_{l^k}(\mathbf{x}_j^k) + \mathbf{p}_{G}(\mathbf{x}_j^k) + \mathbf{p}_{c^k}(\mathbf{x}_j^k)\right] \bigr\}
% \end{equation}

% The above equation represents the relationship between the initial threshold value and prediction values of three models, where $\alpha_t$ is the ratio factor to control the initial threshold value, $U$ is the total number of the unlabeled data. The dynamic relationship between the threshold value and communication rounds will be shown in \ref{eq-threshold}

\subsubsection{Local Model Retraining \& Server Aggregation}
{\name} is able to generate pseudo labels for unlabeled data and automatically add high-quality unlabeled data to the training set. Thus, based on the new training data $\{\mathcal{D}_L^k, \mathcal{D}_{P}^k\}$, we can retrain each local model by minimizing the following loss function:
\begin{equation}
\label{total_loss}{{{L}_{total}}={{L}_l}(\mathcal{D}_L^k)+{\lambda {L}_p}(\mathcal{D}_P^k)},
\end{equation} 
where ${{L}_l}(\mathcal{D}_L^k)$ is the loss on the labeled data calculated by Eq.~(\ref{eq:fedavg}), 
$\lambda$ is a hyperparameter to balance the loss obtained from the pseudo-labeled data, and ${L}_p(\mathcal{D}_P^k)$ is the loss of the pseudo-labeled data and defined as follows:
\begin{equation}\label{eq:pseudo_loss}
L_p(\mathcal{D}_P^k) = \min \left[\frac{1}{N_P^k} \sum_{j=1}^{N_P^k} CE \left(f\left(\mathbf{x}^k_j; \boldsymbol{\omega}^k\right), \hat{y}^k_j\right)\right],
\end{equation}
where $N_P^k$ is the number of selected high-quality pseudo-labeled data, and $\hat{y}^k_j$ is the pseudo label of the unlabeled data $\mathbf{x}^k_j$. 
We maintain the same uploading, model aggregation and downloading methods as in the pre-training stage to retrain the model in the pseudo label learning stage until {\name} converges or runs $T_2$ times. However, the difference is that we train the combined network, i.e., $c^k$, at client side instead of the renewed global model $G$ as FedAvg. The whole learning procedure is shown in Algorithm 1.
%However, when the new global model was lowered to the client, we extracted part of the network parameters of the global model and combined them with the local network parameters saved in the last round to form a new local network. 

\subsubsection{Layers Selection for Model Splicing}
Obviously, the global model generally has the better generalization ability than the local models after aggregation. In contrast, the local models show better performance on their corresponding local datasets due to the difference among local trainsets. For a deep neural network, we can call the first few layers as shallow layers, and the counter-down few layers as deeper layers. The training of deep neural networks is often a process of extracting high-dimensional information from data. The deeper the network layer, the more abstract the information processed. Based on that we believe the shallow layers are tending to focus more on common features of a dataset, while deeper layers for more specific ones. Thus, the combination of the shallow layers of the global model and the deeper layers of local models could creat a stronger combined networks.
 
 In our work, considering that the CNN network used has a relatively simple structure, we choose the first two convolutional networks as shallow layers, and the remaining network structure as deeper layers. Our method can replace the data set and network structure relatively easily. When faced with a complex network structure, in order to achieve the optimal training effect, further experiments are needed to find the optimal network splicing method. But in this work, our experiment results show that appropriate adjustments to shallow layers will not cause a huge difference in classification accuracy. Therefore, in the subsequent experimental sections, we will focus on the method itself instead of the selection of layers structure.

\begin{CJK*}{UTF8}{gkai}
%SetUp
    \begin{algorithm}\label{alg:1}
        \caption{FedTriNet}
        \begin{algorithmic}[1] 
        \Require ${D}_L$ and ${D}_U$
        %\Require ${D}_U$

        \Procedure {Phase I }{\textbf{Pretrain}}
            \State Initialization: ${\omega}_{0}$\algorithmiccomment{initialize weights}
             %for
            \For{each communication round $t = 1,2,3\cdots,T_1$}
                \State $L_{t} = \{l^k\}_{k=1}^{N_t}\gets\mathcal{L} = \{l^k\}_{k=1}^N$\algorithmiccomment{random selection of clients for server aggregation}
                \For{each client $k\in L_{t}$ in parallel}
                        \State $\Delta\omega^k_{t+1}, l^k_{t} \gets \text{Local Update I}(\omega^G_t, D^k_L)$\algorithmiccomment{local model training with labeled data}
                \EndFor
                \State $\omega^G_{t+1}\gets \omega^G_{t} + \frac{1}{N_t}\bigl(\sum_{i=1}^{N_t} \Delta\omega^k_{t+1}\bigr)$\algorithmiccomment{server aggregation by weights averaging}
            \EndFor
        \EndProcedure
        % \State
        \Procedure{Phase II }{\textbf{Pseudo Label Learning}}
        \For{each communication round $t = 1,2,3\cdots,T_2$}
                \State $L_{t} = \{l^k\}_{k=1}^{N_t}\gets\mathcal{L} = \{l^k\}_{k=1}^N$\algorithmiccomment{random clients selection}
                \For{each client $k\in L_{t}$ in parallel}
                        \State $\Delta\omega^k_{t+1}, l^k_{t} \gets \text{Local Update II}(\omega^G_t, \omega^G_{t-1}, D^k_L, D^k_U, t)$\algorithmiccomment{client k's local training and pseudo labeling}
                \EndFor
                \State $\omega^G_{t+1}\gets \omega^G_{t} + \frac{1}{N_t}\bigl(\sum_{i=1}^{N_t} \Delta\omega^k_{t+1}\bigr)$\algorithmiccomment{server aggregation by weights averaging}
            \EndFor
        \EndProcedure
        \State
        \Function {Local Update I }{$\omega^G_t, D^k_L$}
        \For{i in local epochs}
        \For{$B_1$ in $D^k_L$}
            \State $l^k_{t}\gets L_l\bigl(\omega^k_t, B_1\bigr)$\algorithmiccomment{supervised loss computation}
            \State $\omega^k_{t+1}\gets \omega^k_{t} - \eta\nabla l^k_{t}$\algorithmiccomment{mini batch gradient descent}
            \EndFor
        \EndFor
        \State \Return $\omega^k_{t+1}, l^k_t$\algorithmiccomment{return updated weights and client loss}
        \EndFunction
        
        \Function{Local Update II }{$\omega^G_t, \omega^G_{t-1}, D^k_L, D^k_U, t$}
        
        \State $\omega^k_t = \omega^G_{t}[0:n] \cup \omega^k_{t-1}[n+1:m]$\algorithmiccomment{model combination}
        
        \State$\hat{Y}_j^k = \operatorname*{arg\,max} \mathbf{p}_j^k(D^k_U)$\algorithmiccomment{joint prediction}
        
        \If{${p}_j^k(D^k_U) > \theta(t)$}\algorithmiccomment{compare joint prediction value with threshold}
            \State$\hat{D}^k_U \gets \hat{Y}_j^k$\algorithmiccomment{pseudo labeling}
        \EndIf
        \For{i in local epochs}
        \For{$B_1, B_2$ in $D^k_L, \hat{D}^k_U$}
            \State $l^k_{t}\gets L_p\bigl(\omega^k_t, B_1, B_2\bigr)$\algorithmiccomment{pseudo labeled loss computation}
            \State $\omega^k_{t+1}\gets \omega^k_{t} - \eta\nabla l^k_{t}$\algorithmiccomment{mini batch gradient descent}
        \EndFor
        \EndFor
            \State \Return $\omega^k_{t+1}, l^k_t$\algorithmiccomment{return updated weights and client loss}
        \EndFunction

        \end{algorithmic}
    \end{algorithm}
\end{CJK*}
\section{Experiment}\label{sec:experiemts}
In this section, we first introduce the experimental settings, implementation, and then present the experimental results under both IID and Non-IID scenarios.

\subsection{Experimental Settings}

\subsubsection{Datasets}
In our experiments, we use three public datasets in our experiment: MNIST, Fashion-MNIST, and SVHN. For MNIST and Fashion-MNIST datasets, both of them are divided into a training set of 60,000 images and a test set of 10,000 images. For the SVHN dataset, 73,257 digits are used for training and 26,032 digits for testing. The three datasets are all used for the image classification task with 10 categories (i.e., $C=10$).

\subsubsection{Data Distribution Setting}
% For all the three image datasets, each of them will be randomly shuffled and divided into 10 shards for different clients which will have $\frac{D}{N}$ samples separately, where $D$ is the total number of training data and $N$ is the number of clients. We use $\alpha$ to represent the labeled data ratio of all the clients. In other words, there are $\frac{D}{N}*\alpha$ labeled data in every client as well as $\frac{D}{N}*(1-\alpha)$ unlabeled ones. To fully estimate the performance of our model, we use labeled data amount and classes contained in each client to control the data distribution of our federated framework,\jq{I don't quite understand this sentence.} resulting in two distribution settings :

% \subsubsection{IID setting.} Both labeled and unlabeled data of the train set will be shuffled randomly then allocated to each client. Under this condition, every client will have nearly same data distribution either on labeled data amount or classes number.

% \subsubsection{NoIID setting}
% We adapted different division strategy for labeled data and unlabeled data. For unlabeled data distribution, every client contains all categories of data, which is exactly same with the IID condition. While for labeled data distribution, there are only two categories of them stored in each client with a labeled data ratio $\alpha$ to control label amount.
Each of the three datasets is randomly shuffled and divided into 10 shares for $N$ different clients. Given training data number $D$ and labeled data proportion $\alpha$, there is $D \times \alpha /N$ labeled data and $D \times (1 - \alpha) /N$ unlabeled data for each client. To estimate our model performance, we use labeled data amount and class categories in each client to control the data distribution. Furthermore, we consider non-IID and IID distribution settings respectively. For the IID setting, both labeled and unlabeled data in the train set will be shuffled randomly and allocated to each client. For non-IID setting, every client owns all categories of unlabeled data and only two categories of labeled data.

% \smallskip
\subsubsection{Baselines}
To fairly validate the proposed {\name} framework, we use one federated supervised learning model {\texttt{FedAvg}}~\cite{McMahan2017CommunicationEfficientLO}, and two federated semi-supervised learning models, which are {\texttt{FedSem}}~\cite{Albaseer2020ExploitingUD} and {\texttt{FedMatch}}~\cite{Jeong2020FederatedSL}. 
\begin{itemize}
    \item \underline{\texttt{FedAvg}}~\cite{McMahan2017CommunicationEfficientLO}, proposed by McMahan, et al., presents how to conduct federated learning of deep networks with decentralized data based on iterative model averaging under the communication cost constraints. Each client updates local models by stochastic gradient descent and the server performs model averaging. With empirical evaluation, this approach is robust to unbalanced and non-IID data distributions.
    
    \item \underline{{\texttt{FedSem}}}~\cite{Albaseer2020ExploitingUD}, proposed by Abdullatif Albaseer, et al., combines pseudo labeling idea with federated semi-supervised learning problems in the smart city application. In this work, the training process is divided into two phases. In phase one, with the existing labeled data to supervise the training process, the local model obtains the certain classification ability. With the model, the local unlabeled data is labeled with the predicted value as a pseudo-label. In phase two, the whole federated framework will continue the same training process as in Phase I with data with real labels and pseudo-labels..
    
    \item \underline{{\texttt{FedMatch}}}~\cite{Jeong2020FederatedSL} adopts the idea of consistency regularization and designs two kinds of loss functions to guide the training of the model, namely Inter-client Consistency Loss and Data-level Consistency Regularization. The idea of consistency regularization is that the output of the predictor is expected to be as consistent as possible between an original sample and the processed version by data enhancement (the idea of consistency). In the process of server parameter delegation, in addition to the original model of the client, several models of other clients will be sent to the client as helper agents. The final purpose of local unsupervised training is to minimize the difference between the prediction results of the local model and the labels provided by each consensus model as small as possible.
\end{itemize}
In the following subsections, we will compare the performance of our model {\name} with the discussed baselines under two different data distribution settings.

\subsection{Implementation}
%\jq{This paragraph has been modified within the content rather than creating new paragraphs.}

When implementing all baselines and {\name}, we use the same local model for each client. A Convolutional Neural Network (CNN) is used for the image classification tasks of three datasets. 
We adopt the weak data argumentation technique on the three datasets for all the baselines and {\name}, where the main process contains random reflect, flip, contrast adjustment, grayscale, and crop. For all the IID experiments, we set the local training epochs as $5$ and total communication rounds $T$ as $100$. For the non-IID setting, the local training epoch is set to the same number as IID along with other parameters. For MNIST, the pre-training stage rounds $T_1$ is $40$ and pseudo label learning stage rounds $T_2$ is $60$; for Fashion-MNIST,  $T_1$ is $30$ and $T_2$ is $70$; for SVHN, $T_1$ is $60$ and $T_2$ is $40$. Besides, the client number is fixed as 10. The local training batch size is set as 50 for both labeled data and unlabeled data.

\begin{table*}[!t]
\centering
%\vspace{-0.05in}
\caption{Accuracy on the three datasets under the IID setting, where all clients have the same distribution.}
\begin{tabular}{c|ccc|ccc|ccc}
\toprule
\textbf{Dataset}             & \multicolumn{3}{c|}{\textbf{MNIST}} & \multicolumn{3}{c|}{\textbf{Fashion-MNIST}} & \multicolumn{3}{c}{\textbf{SVHN}} \\ \hline
\textbf{\# Labeled Data} & 60         & 600        & 6000      & 600           & 3000         & 6000         & 1000       & 3000      & 6000      \\ 
\midrule
\texttt{FedAvg}     & 29.26\%    & 88.26\%    & 96.46\%   & 65.19\%       & 74.54\%      & 78.32\%      & 27.82\%    & 78.44\%   & 87.77\%   \\ \hline
\texttt{Fedsem}     & 39.49\%    & 84.54\%    & 96.31\%   & 45.33\%       & 74.29\%      & 78.78\%      &     18.47\%       &      76.22\%     &      86.16\%     \\ \hline
\texttt{FedMatch}  &     46.75\%       &      89.28\%      &       97.14\%    &       69.56\%        &      77.28\%        &      79.15\%        &     59.61\%       &     78.94\%      &     88.26\%      \\ \hline
\texttt{{FedTriNet}}  & \textbf{77.25\%}    & \textbf{93.80\%}    & \textbf{97.56\%}   & \textbf{71.88\% }      & \textbf{78.01\%}      & \textbf{81.00\%}      & \textbf{63.99\%}    & \textbf{79.47\%}   & \textbf{89.48\% }  \\ 
\bottomrule
\end{tabular}

\label{tab-iid}
\end{table*}

\subsection{Performance Evaluation for the IID Setting}
Table~\ref{tab-iid} shows the performance of all the approaches under the IID scenario.
From Table~\ref{tab-iid}, we can observe that {\name} shows the best performance with all the given settings on the three datasets. Besides, with the increase of the number of labeled data, the performance of all the approaches increases. 
Although \texttt{Fedsem} also uses two-phase training, it cannot even outperform the supervised method \texttt{FedAvg}. The reason is that after phase I training, \texttt{Fedsem} generates pseudo labels for all the unlabeled data, which are then used for phase II training. Since the quality of pseudo labels is pretty low when the number of labeled data is small, the misleading information further hurts the learning of Phase II. Thus, \texttt{Fedsem} performs worst compared with other baselines.
\texttt{FedMatch} uses data augmentation, inter-client consistency, and disjoint learning techniques to achieve the second-best performance for all the settings. 
%It shows best robustness to the labeled data amount\jq{how we can say the best stability and robustness? any supportive findings} among three baseline methods \textcolor{red}{
It has $46.75\%$ accuracy on the MNIST dataset when the number of labeled data is set to 60 for all the clients, while {\verb|FedAvg|} and {\verb|Fedsem|} collapse. For the experiments on the SVHN dataset, we can also see that the {\verb|FedMatch|} reaches $59.61\%$ compared to the poor performance of both {\verb|FedAvg|} and {\verb|Fedsem|}.
%}

% With the decrease of the labeled data amount, the performance of {\verb|FedAvg|} and {\verb|Fedsem|} drops into a significantly low level comparing to {\verb|FedMatch|} and our proposed method. This is because the limited  supervised information makes the supervised learning methods have difficulty in maintaining the stable model training process. For instance, given 60 labeled data from MNIST will make the training inefficient and unstable. From here, each labeled data will cause significant impact on the supervised training result, and the extreme similarity between the images may cause the poor performance. As shown in \ref{Fig:mnist sample}, Figure 8 and 9 from MNIST could be visually confusing to each other due to the writing styles of human, which also happens on other figures like 7 and 9. Once a client's local training dataset has 2 visually similar pictures out of 6 but different labels, the corresponding model performance could be unpleasant \jq{to here, I may not understand what is talked about}.

% \begin{figure}[htbp]  
% \centering  
% \includegraphics[height=3cm, width=7cm]{samples/Figure/mnist sample.pdf}  
% \caption{The proposed {\name} Framework} 
% \label{Fig:mnist sample}
% \end{figure}

\subsection{Performance Evaluation for the Non-IID Setting}
In Table~\ref{tab-noniid}, with the Non-IID setting, our proposed approach {\name} still outperforms all the baselines. Especially for the experiment with $6000$ labeled data, the accuracy of {\verb|FedTriNet|} was $3.21\%$ and $16.96\%$ higher than that of {\verb|FedAvg|} on MNIST and SVHN, respectively. Compared with the results listed in Table~\ref{tab-iid}, we find that all the accuracy drops. This observation is in accord with the fact, that is, the Non-IID setting is more challenging than the IID setting for federated learning due to the data and label imbalance.

% What worth to mention is that, {\verb|Fedsem|} shows extreme discomfort with NonIID data, which has the greatest drop on performance among all the methods. This phenomenon further prove the fragility of the traditional self-training based pseudo-labeling method. While {\verb|FedTriNet|} depending  on the three-player framework achieves better ability for the heterogeneity challenge, even has higher classification accuracy for SVHN 3000 labeled data setting than IID one.

It is worth mentioning that {\verb|Fedsem|} shows extreme discomfort with Non-IID data, which has the greatest drop in performance among all the methods. This phenomenon further proves the fragility of the traditional self-training based pseudo-labeling method. {\verb|FedTriNet|} with the three-player framework achieves the better ability for the heterogeneity challenge and even achieves higher classification accuracy for SVHN 3000 labeled data setting than the IID one.

\begin{table*}[!t]
\centering
\caption{Accuracy on the three datasets under the non-IID setting.}
\begin{tabular}{c|ccc|ccc|ccc}
\toprule
\textbf{Dataset}             & \multicolumn{3}{c|}{\textbf{MNIST}} & \multicolumn{3}{c|}{\textbf{Fashion-MNIST}} & \multicolumn{3}{c}{\textbf{SVHN}} \\ \hline
\textbf{\# Labeled Data} & 60         & 600        & 6000      & 600           & 3000         & 6000         & 1000       & 3000      & 6000      \\ 
\midrule
\texttt{{FedAvg}}     & 26.29\%    & 77.67\%    & 91.79\%   & 63.86\%       & 70.56\%      & 74.28\%      & 19.38\%    & 46.70\%   & 66.45\%   \\ \hline
\texttt{{Fedsem}}     & 32.06\%    & 74.12\%    & 84.11\%   & 17.06\%       & 56.67\%      & 63.93\%      & 18.95\%    & 49.02\%   & 53.94\%   \\ \hline
\texttt{{FedMatch}}   &     69.28\%       &      79.15\%      &      93.20\%     &      65.44\%         &      71.26\%        &      74.81\%        &      54.34\%      &      74.27\%     &      79.42\%     \\ \hline
\texttt{{FedTriNet}}  & \textbf{79.60\%}    & \textbf{82.55\%}    & \textbf{95.00\%}   & \textbf{69.50\% }      & \textbf{72.77\%}      & \textbf{75.05\%}      &      \textbf{57.26\%}      & \textbf{82.78\%}   & \textbf{83.41\%}   \\ 
\bottomrule
\end{tabular}

\label{tab-noniid}
\end{table*}
%\jq{please highlight the values (bold) you would like to discuss in the table}
% \begin{table}
% \centering
% \vspace{0.05in}
% \begin{small}
% \begin{tabulary}{\linewidth}{l|CCC}
% \toprule
% \textbf{Model} & \textbf{MNIST} & \textbf{FMNSIT} & \textbf{SVHN} \\
% \midrule
% FedAvg      & 88.81\%   & 74.13\% & 69.43\% \\
% FedSem      & 84.18\%   & 73.41\% & 70.70\% \\
% FedMatch  & 88.34\%  & 74.57\% & 71.04\% \\
% \midrule
% FedTriNet  & \textbf{93.51\%} & \textbf{75.45\%} & \textbf{73.38\%}  \\

% \bottomrule
% \end{tabulary}
% \vspace{0.1in}
% % \caption{Accuracy on the three datasets under the NonIID setting}
% \label{tab-iid}
% \end{small}
% \vspace{-0.2in}
% \end{table}

\subsection{Ablation Study}
In this experiment, we aim to conduct the model insight analysis removing each of the following modules in {\name}, and the results are shown in Table~\ref{tab-tricks}.
\begin{itemize}
    \item \emph{Threshold Guarantee Mechanism.} In the later stage of model training, the threshold of pseudo-labeling control is maintained at a relatively high value, which ensures that the pseudo-labeling data updated into the training set has consistently high quality and does not affect the model performance. The experiment results show that the shutdown of the Threshold Guarantee Mechanism will cause the performance drop by a few percent. The higher the original accuracy is, the less the drop is. This indicates that this mechanism can maintain the pseudo label quality to avoid introducing misleading information.
    
\begin{table*}[!h]
\centering
\caption{Ablation experiment on the three datasets under the IID setting.}
\begin{tabular}{c|c|c|c}
\toprule
\textbf{Dataset}                                    & \textbf{MNIST}               & \textbf{Fashion-MNIST}       & \textbf{SVHN} \\\hline
\multicolumn{1}{c|}{\textbf{\# Labeled Data}}   & \multicolumn{1}{c|}{600}     & \multicolumn{1}{c|}{1000}    & 3000          \\ \midrule
\multicolumn{1}{l|}{\name}                      & \multicolumn{1}{c|}{\textbf{93.80\%}} & \multicolumn{1}{c|}{\textbf{78.07\%}} & \textbf{79.47\%}       \\ \hline
\multicolumn{1}{l|}{-Threshold Guarantee Protection} & \multicolumn{1}{c|}{90.22\%} & \multicolumn{1}{c|}{71.70\%}        &       74.93\%        \\ \hline
% \multicolumn{1}{c|}{Pseudo Labeler Selection}       & \multicolumn{1}{c|}{\textbf{93.94\%}} & \multicolumn{1}{c|}{72.37\%}        &       69.13\%        \\ \hline
\multicolumn{1}{l|}{-Fine-tuning}                      & \multicolumn{1}{c|}{87.20\%} & \multicolumn{1}{c|}{72.34\%}   &            71.30\%   \\\hline
\multicolumn{1}{l|}{-Pseudo Labeling}                      & \multicolumn{1}{c|}{86.50\%} & \multicolumn{1}{c|}{70.75\%}   &         72.73\%      \\
\bottomrule
\end{tabular}
\label{tab-tricks}
\end{table*}
    
    \item \emph{Fine-tuning.} In order to make the network parameters more suitable for local data, a new network constructed from the first several layers of the new global model and the several latter layers of the old local model will be labeled with fine-tuning operation with the labeled data. Another important reason for the fine-tuning process is that if the labeled data and unlabeled data are trained with the same model parameter respectively (that is, the model parameters are shared), the unlabeled training process may cause the model to forget the knowledge learned from the labeled data. To make a fair comparison, we compensate for additional local training epochs in the latter half of the communication round during the training process of the baseline models. From the results, we can see that the fine-tuning operation is essential for model learning.
    
    \item \emph{Pseudo Labeling.} In the proposed {\name}, we first pre-train the model and then conduct the pseudo label learning. During the second phase, we remove the pseudo labeling operation and directly use the combined network $c^k$ and the labeled data to train the model. We can observe that the performance of pseudo labeling significantly drops compared with that of {\name}. These results clearly demonstrate the importance of pseudo labeling for federated semi-supervised learning.
\end{itemize}

\subsection{Phase Round Combination}
For a constant total communication round setting, different phase I and phase II ratios may cause different model performances. Due to the different amounts of information in different images, a properly supervised learning training period would benefit the model performance more than an early entrance to the semi-supervised phase, a.k.a. pseudo label stage. 
Table~\ref{tab-roundcombine-iid} demonstrates the performance changes with the different phase rounds under both IID and Non-IID settings. Here the first column represents the different phase round combinations. For example, $30+70$ means the experiment is 30 phase I rounds and 70 phase II rounds.

\begin{figure*}[htbp]  
\centering  
\includegraphics[height=12cm, width=16.4cm]{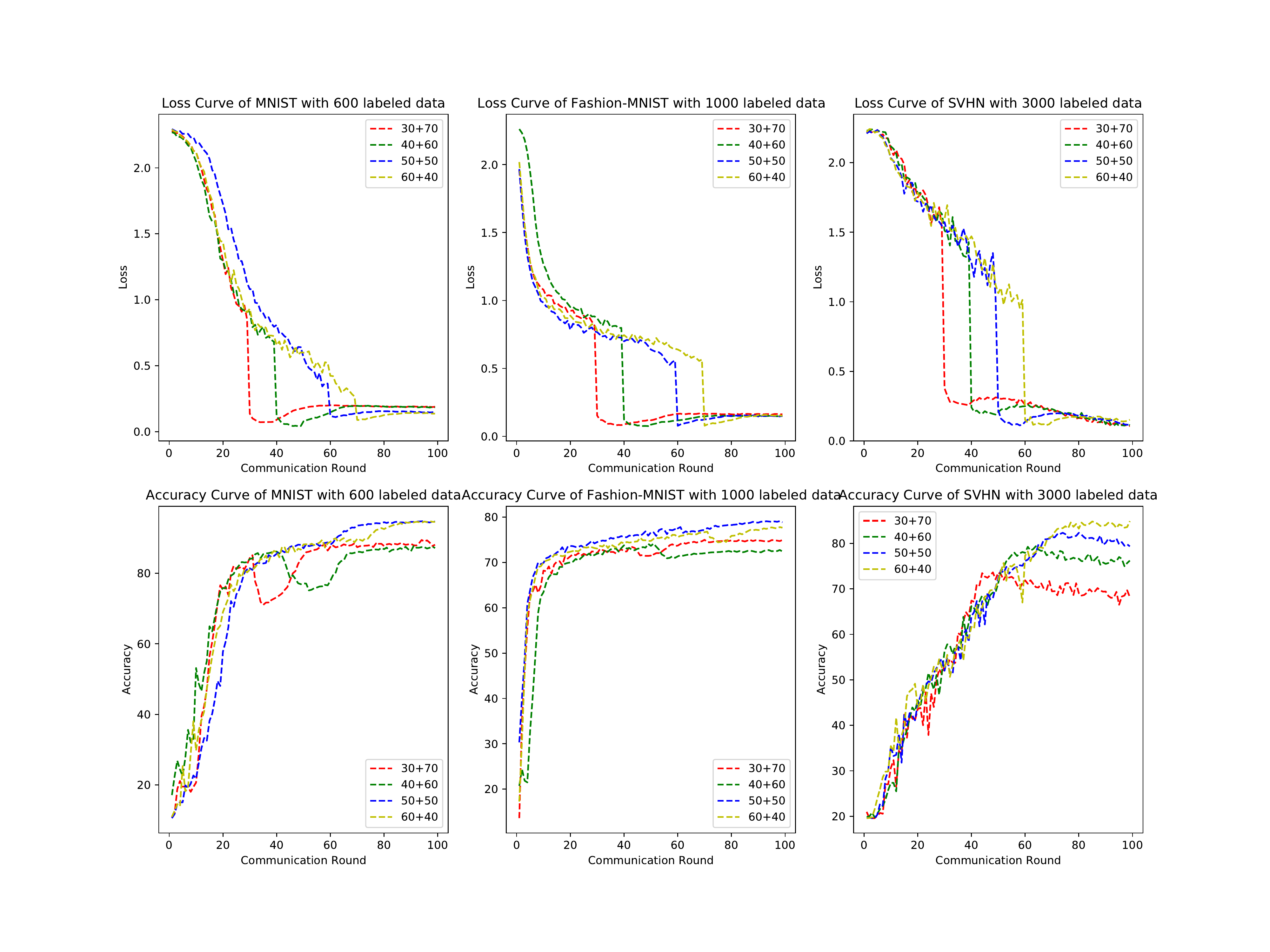}  
\caption{Phase Round Combination Results under IID setting} 
\label{Fig:Phase Round Combination iid}
\end{figure*}

\begin{figure*}[htbp]  
\centering  
\includegraphics[height=12cm, width=16.4cm]{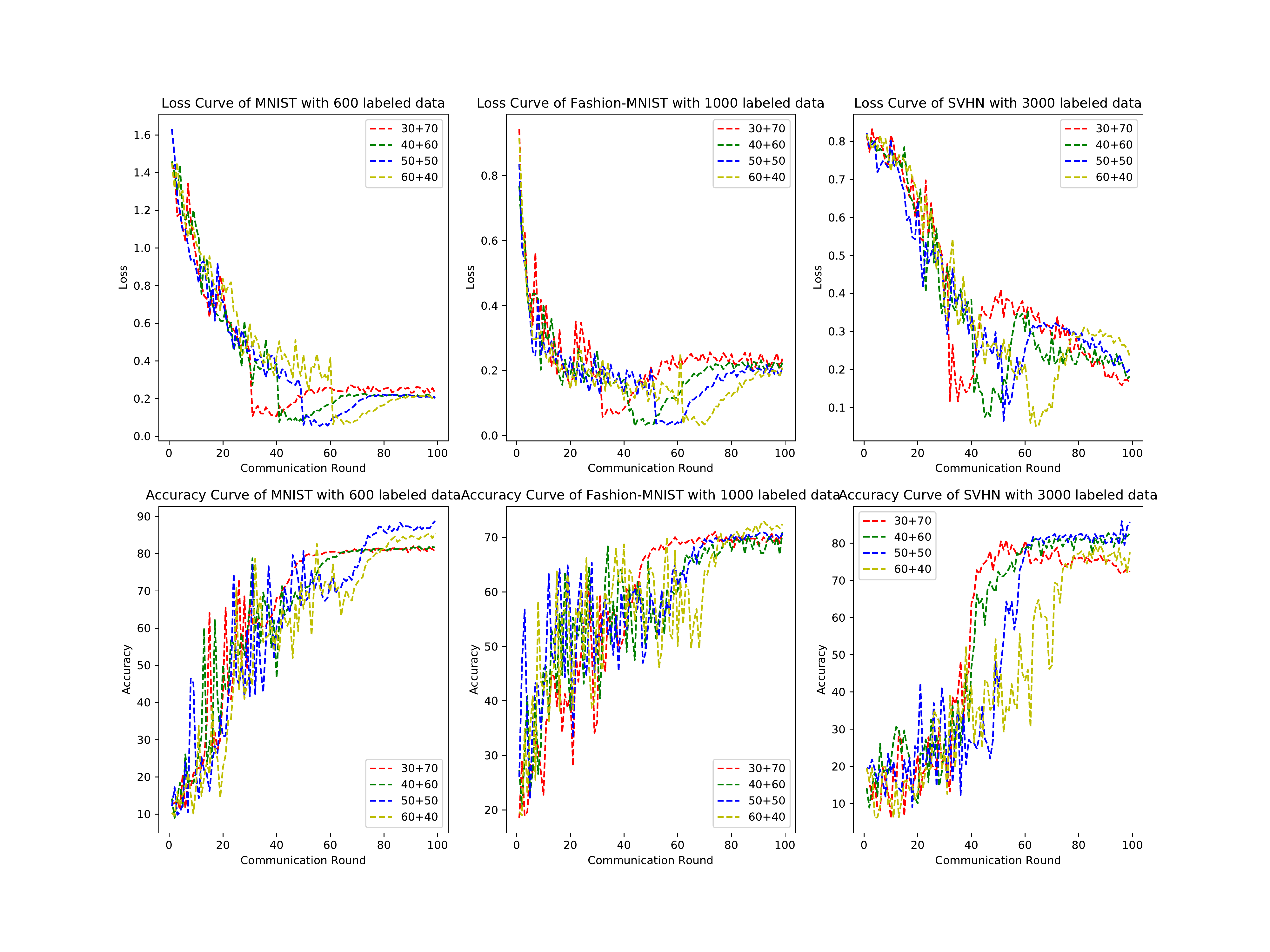}  
\caption{Phase Round Combination Results under Non-IID setting} 
\label{Fig:Phase Round Combination noniid}
\end{figure*}

Our IID experiments on MNIST, Fashion-MNIST, and SVHN show that under the parameter controlling condition, the model's performance will increase as the supervised learning round increases till reaching a peak and then decrease. In our experiments, the most proper ratio for MNIST and Fashion-MNIST is 50 rounds in phase I and 50 rounds in phase II (50+50). For SVHN, the setting is 60+40 rounds.
While for the non-IID setting, there is no apparent accuracy changing trends with the phase round combination for all three datasets. The best results of the three datasets MNIST, Fashion-MNIST, and SVHN are achieved with the settings of 50+50 rounds, 60+40 rounds, and 50+50 rounds, respectively. The overall results indicate the robustness of our algorithm for data distribution.

Fig \ref{Fig:Phase Round Combination iid} illustrates the loss and accuracy curves of phase round combination experiment for three datasets under IID setting. In loss curves, the start round of phase II usually causes a plummet, which is because the pseudo-labeled data is added into the training set. This results in the accuracy fluctuation within a narrow range, which usually happens to the curves whose finally performance is not satisfying, either. A proper phase round combination will allow the model to avoid introducing too many pseudo labeling errors into the training. For instance, in the accuracy curve of SVHN, the 60+40 case achieves best result, while the 30+70 one does not rise but fall. What noticeable else, is in phase II, the loss curves firstly increase then decrease, which reflects the  correction function of our methods. Fig \ref{Fig:Phase Round Combination noniid} shows the similar phenomenons under NonIID setting, with larger training curve fluctuation.

% \jq{Modified}The further experiment that forbids the training set update with all the other training tricks reserved has shown that our pseudo label method indeed significantly improves the model performance. The experiment result has an average drop of 10 percent for MNIST classification accuracy and 3 percent for Fashion-MNIST.

% \usepackage{booktabs}
\begin{table*}[!ht]
\caption{Accuracy on the three datasets under different phase I and phase II rounds combination for IID and Non-IID data.}
\centering
\vspace{0.05in}
\begin{tabular}{c|c|c|c||c|c|c}
\toprule
\textbf{Setting} & \multicolumn{3}{c||}{IID} & \multicolumn{3}{c}{Non-IID}\\\hline
\textbf{Dataset}             & \textbf{MNIST} & \textbf{Fashion-MNIST} & \textbf{SVHN} & \textbf{MNIST} & \textbf{Fashion-MNIST} & \textbf{SVHN} \\\hline 
\textbf{\# Labeled Data} & 600            & 1000                   & 3000     & 600            & 1000                   & 3000     \\\midrule
30+70                        & 88.56\%        & 74.79\%                &      68.14\%   & 80.94\%        &     71.03\%                   &   72.31\%      \\ \hline
40+60                        &  94.66\%        & 75.84\%                &      78.44\%     & 80.65\%        &          67.46\%              &    82.36\%   \\ \hline
50+50                        & \textbf{94.80\%}        & \textbf{79.05\%}                &         77.45\%    & \textbf{87.06\%}        &         70.78\%               & \textbf{84.57\%}   \\ \hline
60+40                        & 94.64\%        & 77.67\%                & \textbf{83.15\%}     &   85.51\%             &           \textbf{72.45\%}             & 77.57\%  \\ \bottomrule
\end{tabular}

\label{tab-roundcombine-iid}
\end{table*}

% \usepackage{booktabs}
% \begin{table}[]
% \begin{tabular}{@{}c|c|c|c}
% \toprule

% \textbf{Dataset}             & \textbf{MNIST} & \textbf{Fashion-MNIST} & \textbf{SVHN} \\\hline
% \textbf{Labeled Data Number} & 600            & 1000                   & 3000          \\ \midrule
% 30+70                        & 86.94\%        &     71.03\%                   &   72.31\%            \\ \hline
% 40+60                        & 81.65\%        &          67.46\%              & 80.45\%       \\\hline
% 50+50                        & 87.06\%        &         70.78\%               & 79.36\%       \\ \hline
% 60+40                        &   85.51\%             &           72.45\%             & 77.57\%       \\ \bottomrule
% \end{tabular}
% \caption{Accuracy on the three datasets under different phase I and phase II rounds combination for NonIID data}
% \label{tab-roundcombine-noniid}
% \end{table}

\section{Conclusion}\label{sec:con}
Federated learning is a new collaborative learning approach without sharing client data and can apply to many real-world applications. Although many federated learning approaches are proposed, they mainly focus on the supervised setting, which is not realistic due to the strict requirement that all the client data have corresponding labels. Only a few studies are trying to explore the power of unlabeled data, but they either need to know the information of neighboring clients or introduce low-quality pseudo labels into the model training. 

To address these problems, in this paper, we propose an effective pseudo labeling method with three players for federated semi-supervised learning called {\name}. {\name} consists of two learning phases. In the first phase, we use the labeled data to pre-train {\name} using FedAvg. In the second phase, we aim to use unlabeled data by generating high-quality pseudo labels. Towards this end, we propose to use three networks, including one local model, one global model, and one combined model from the previous two models. Besides, a quality control mechanism is proposed to generate a global-level threshold, which dynamically changes with the global training rounds. The corresponding unlabeled data can be added to the training set only when the maximum probability value is larger than this threshold. Finally, {\name} retrains the combined model with the new training data. We conduct experiments on three benchmark datasets to show the effectiveness of the proposed {\name} compared with state-of-the-art baselines.

\bibliographystyle{IEEEtran}
\bibliography{ref}

\end{document}